\documentclass[sigconf]{acmart}
\AtBeginDocument{%
  }

\copyrightyear{2025}
\acmYear{2025}
\setcopyright{cc}
\setcctype{by}
\acmConference[MM '25]{Proceedings of the 33rd ACM International Conference on Multimedia}{October 27--31, 2025}{Dublin, Ireland}
\acmBooktitle{Proceedings of the 33rd ACM International Conference on Multimedia (MM '25), October 27--31, 2025, Dublin, Ireland}\acmDOI{10.1145/3746027.3761996}
\acmISBN{979-8-4007-2035-2/2025/10}

\begin{document}

\title{AdaDocVQA: Adaptive Framework for Long Document Visual Question Answering in Low-Resource Settings}

\author{Haoxuan Li}
\authornote{Both authors contributed equally to this research.}
\email{li-hx24@mails.tsinghua.edu.cn}
\affiliation{%
  \institution{Tsinghua Shenzhen International Graduate School, Tsinghua University,}
  \city{Shenzhen}
  \state{Guangdong}
  \country{China}
}

\author{Wei Song}
\authornotemark[1]
\email{sonwe@mail2.gdut.edu.cn}
\affiliation{%
  \institution{School of Automation, Guangdong University of Technology,}
  \city{Guangzhou}
  \state{Guangdong}
  \country{China}
}

\author{Aofan Liu}
\authornotemark[1]
\email{af.liu@stu.pku.edu.cn}
\affiliation{%
  \institution{School of Information Engineering, Peking University,}
  \city{Shenzhen}
  \state{Guangdong}
  \country{China}
}

\author{Peiwu Qin}
\authornote{Corresponding author.}
\email{pwqin@sz.tsinghua.edu.cn}
\affiliation{%
  \institution{Tsinghua Shenzhen International Graduate School, Tsinghua University,}
  \city{Shenzhen}
  \state{Guangdong}
  \country{China}
}

\renewcommand{\shortauthors}{Haoxuan Li, Wei Song, Aofan Liu, and Peiwu Qin}

\begin{abstract}
    Document Visual Question Answering (Document VQA) faces significant challenges when processing long documents in low-resource environments due to context limitations and insufficient training data. This paper presents AdaDocVQA, a unified adaptive framework addressing these challenges through three core innovations: a hybrid text retrieval architecture for effective document segmentation, an intelligent data augmentation pipeline that automatically generates high-quality reasoning question-answer pairs with multi-level verification, and adaptive ensemble inference with dynamic configuration generation and early stopping mechanisms. Experiments on Japanese document VQA benchmarks demonstrate substantial improvements with 83.04\% accuracy on Yes/No questions, 52.66\% on factual questions, and 44.12\% on numerical questions in JDocQA, and 59\% accuracy on LAVA dataset. Ablation studies confirm meaningful contributions from each component, and our framework establishes new state-of-the-art results for Japanese document VQA while providing a scalable foundation for other low-resource languages and specialized domains. Our code available at: https://github.com/Haoxuanli-Thu/AdaDocVQA

\end{abstract}




\maketitle

\section{Introduction}
Document Visual Question Answering (Document VQA) stands as a core task in multimodal understanding, requiring machines to comprehend complex document layouts and textual content to accurately answer related questions. Despite significant advances in large-scale vision-language models in this domain, substantial challenges persist when processing long documents and low-resource languages \cite{zhong2024opportunities}. Current methods are primarily constrained by two critical bottlenecks: first, context window limitations of existing models prevent processing of complete documents containing dozens or hundreds of pages; second, for data-scarce languages and domains, the lack of high-quality training data severely limits model performance.

To address these challenges, the Retrieval-Augmented Generation (RAG) paradigm offers new perspectives for long document processing. However, applying RAG to document visual question answering presents numerous technical difficulties: how to effectively retrieve relevant document segments while maintaining semantic integrity, how to optimize text retrieval performance for specific languages, and how to achieve balance between retrieval quality and computational efficiency. Meanwhile, regarding insufficient training data in low-resource scenarios, existing data augmentation methods primarily rely on simple data transformations or back-translation techniques, making it difficult to generate high-quality question-answer pairs with reasoning complexity.

This research proposes AdaDocVQA, a unified adaptive document visual question answering framework for low-resource scenarios. The framework addresses the aforementioned challenges through three core innovations: first, designing a hybrid text retrieval engine that combines lexical matching and semantic similarity with dynamic result selection mechanisms to optimize retrieval quality; second, constructing an intelligent data generation and quality control pipeline capable of automatically generating high-quality reasoning question-answer pairs with multi-level verification; finally, developing an adaptive ensemble inference strategy that improves prediction robustness and efficiency through dynamic configuration generation and early stopping mechanisms.

We validate the framework's effectiveness on Japanese document VQA tasks, which exhibit typical low-resource characteristics: scarce training data, strong language specificity, and complex document structures. Experimental results demonstrate that AdaDocVQA significantly improves long document understanding accuracy while maintaining high efficiency. More importantly, our framework exhibits scalability and can be conveniently adapted to other low-resource languages and specialized domains.
 The main contributions of our work are summarized as follows:
\begin{itemize}
\item  We construct a large-scale Japanese document VQA augmented dataset, introducing automated answer feasibility analysis mechanisms and five types of reasoning question generation strategies, ensuring data quality through multi-level quality verification.
\item We propose a hybrid text retrieval architecture combining TF-IDF with semantic encoding and optimized for specific languages; design an intelligent data augmentation pipeline enabling automatic generation of reasoning questions; develop an adaptive ensemble inference framework that enhances robustness through dynamic configuration generation and early stopping mechanisms.

\item We conduct extensive experiments on the LAVA and JDocQA datasets, where our proposed method achieves state-of-the-art performance, demonstrating its effectiveness and superiority over existing approaches.
\end{itemize}

\section{Related Work}
\subsection{Retrieval-Augmented Generation}
Retrieval-Augmented Generation (RAG) was first introduced by Lewis et al. \cite{lewis2020retrieval} to enhance LLM performance by integrating external knowledge retrieval with text generation. Early approaches followed simple retrieve-and-read paradigms \cite{khandelwal2019generalization}, evolving into naive and advanced RAG systems that incorporated sophisticated retrievers, rerankers, and filters to improve retrieval quality and generation accuracy \cite{cheng2021unitedqa,jiang2023hykge,yoran2023making}.

Recent developments have focused on adaptive RAG strategies enabling LLMs to determine when and what to retrieve. FLARE \cite{jiang2023active} predicts low-confidence tokens to trigger retrieval, while DRAGIN \cite{su2024dragin} leverages uncertainty based on self-attention weights. Adaptive-RAG \cite{jeong2024adaptive} uses smaller LLMs as classifiers for query complexity assessment. Self-RAG \cite{asai2023self} enhances factuality through self-reflection mechanisms, and reasoning-enhanced approaches like ReAct \cite{yao2023react} have integrated step-by-step planning capabilities with external tool usage.
\subsection{Large Vision Language Models}
Recent advances in large vision-language models have demonstrated remarkable progress in multimodal understanding tasks. The LLaVA series \cite{liu2023visual,liu2024improved,li2024llava} achieved precise image-text matching by connecting CLIP \cite{radford2021learning} vision encoders with LLaMA \cite{touvron2023llama} language models through end-to-end visual instruction tuning. LLaVA-NeXT \cite{li2024llava} expanded capacity to 34 billion parameters with 4× pixel resolution support and multitask joint training capabilities. The Qwen-VL series \cite{bai2023qwen,wang2024qwen2,bai2025qwen2} introduced textual encoding strategies for bounding boxes, enabling spatial position awareness through extensive text labeling, with Qwen2.5-VL \cite{bai2025qwen2} demonstrating superior performance in conversational scenarios. InternVL series \cite{chen2024internvl, chen2024expanding, zhu2025internvl3} made significant contributions through progressive scaling strategies, where InternVL constructed a 6-billion parameter vision encoder achieving parameter balance between vision and language branches, while InternVL-2.5 \cite{chen2024expanding} introduced dynamic resolution adaptation supporting multi-scale input from 224 to 1024 pixels through feature pyramid networks. These models collectively represent the current frontier in multimodal AI, demonstrating superior capabilities in fine-grained semantic alignment, zero-shot transfer learning, and complex visual reasoning tasks compared to traditional approaches like CLIP \cite{radford2021learning} and BLIP-2 \cite{li2023blip} .
\subsection{Visual Question Answering}
Document visual question answering has emerged as a challenging task requiring joint understanding of textual and visual elements. Early English datasets such as DocVQA \cite{mathew2021docvqa}, OCR-VQA \cite{mishra2019ocr}, and InfographicVQA \cite{mathew2022infographicvqa} established foundations for single-image document understanding. Recent advances have addressed multi-page comprehension through datasets like MP-DocVQA \cite{tito2023hierarchical} and SlideVQA \cite{tanaka2023slidevqa}, enabling more complex reasoning across document collections. However, the field remains predominantly English-centric, creating significant gaps for other languages. Japanese language understanding has been explored through text-only benchmarks such as JGLUE \cite{kurihara2022jglue}, but these lack comprehensive document-level visual reasoning capabilities. JDocQA \cite{onami2024jdocqa} represents a significant contribution by providing the first large-scale Japanese document VQA dataset, featuring 11,600 question-answer pairs across diverse document types. The dataset's inclusion of unanswerable questions addresses practical deployment concerns by helping models recognize when information is unavailable, thereby reducing hallucination in generative language models.

\section{Method}
\subsection{Framework Overview}

\begin{figure}[h]
  \centering
  \includegraphics[width=\linewidth]{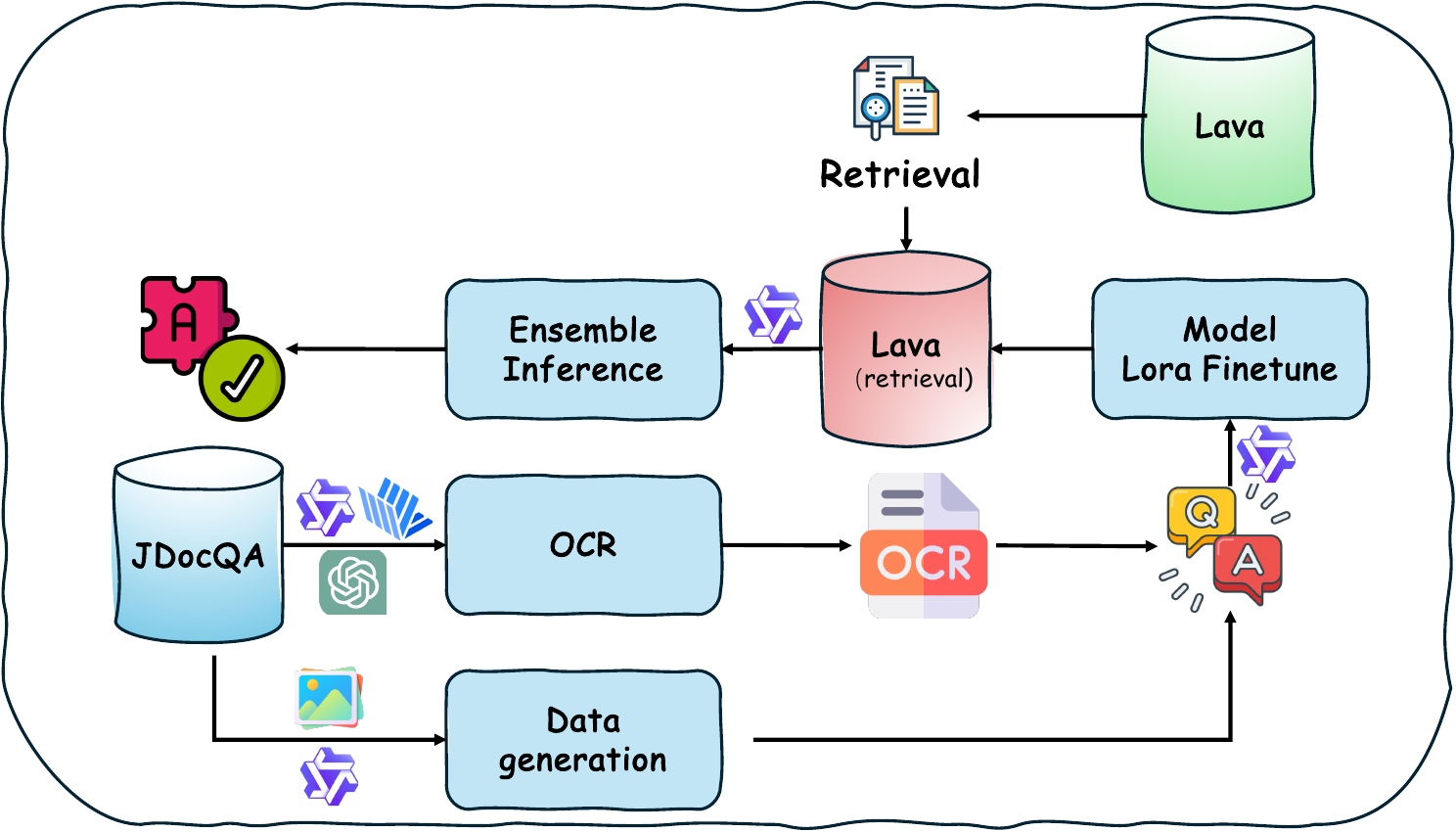}
  \caption{Overview of AdaDocVQA}
  \label{fig:AdaDocVQA}
\end{figure}

\begin{figure}[h]
  \centering
  \includegraphics[width=\linewidth]{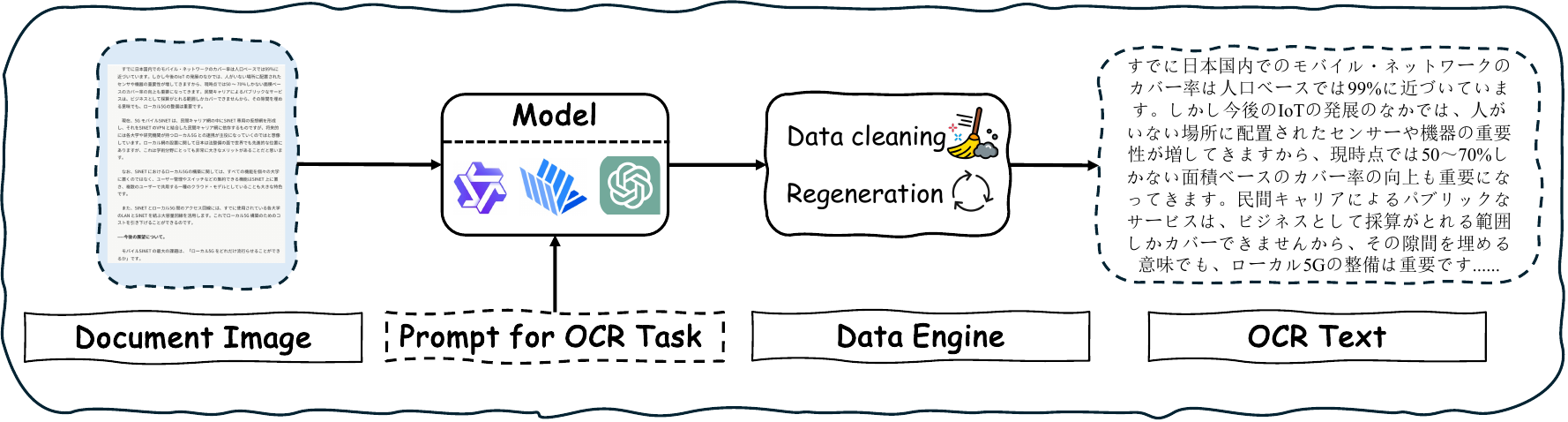}
  \caption{VLM Based OCR System Architecture with Data Cleaning and Regeneration Pipeline}
  \label{fig:OCR}
\end{figure}
Our AdaDocVQA framework operates through an integrated pipeline comprising five core components: enhanced OCR processing with data quality control, intelligent data augmentation, hybrid text retrieval, model fine-tuning, and ensemble inference, as illustrated in Figure \ref{fig:AdaDocVQA}. The system tackles the challenges of long document Visual Question Answering (VQA) through a systematic two-stage approach. In the preprocessing stage, it extracts and enhances textual content from document images, subsequently generating high-quality training data via automated question generation coupled with rigorous quality validation. During inference, the framework employs a hybrid retrieval mechanism to identify contextually relevant document segments, which are then processed by fine-tuned models that leverage adaptive ensemble strategies to deliver robust and accurate predictions.
\subsection{Enhanced OCR and Data Quality Control}
\subsubsection{\textbf{Stage 1: Content-Aware OCR Enhancement}}

Traditional OCR extraction often produces fragmented and context-poor text representations inadequate for complex reasoning tasks. Our enhancement process leverages large vision-language models to extract semantically rich, structurally coherent text representations through specialized prompting strategies. The process, depicted in Figure \ref{fig:OCR}.

The system first loads document images and employs specially designed prompts to guide structured text extraction. The prompts explicitly require maintaining hierarchical structure, describing table and chart relationships, preserving numerical precision, and converting visual elements to textual descriptions. Model outputs undergo post-processing to standardize format while maintaining the document's original logical structure. Enhanced OCR results include complete paragraph structures, table data relationship descriptions, chart trend descriptions, and accurate numerical information.

\begin{figure}[h]
  \centering
  \includegraphics[width=\linewidth]{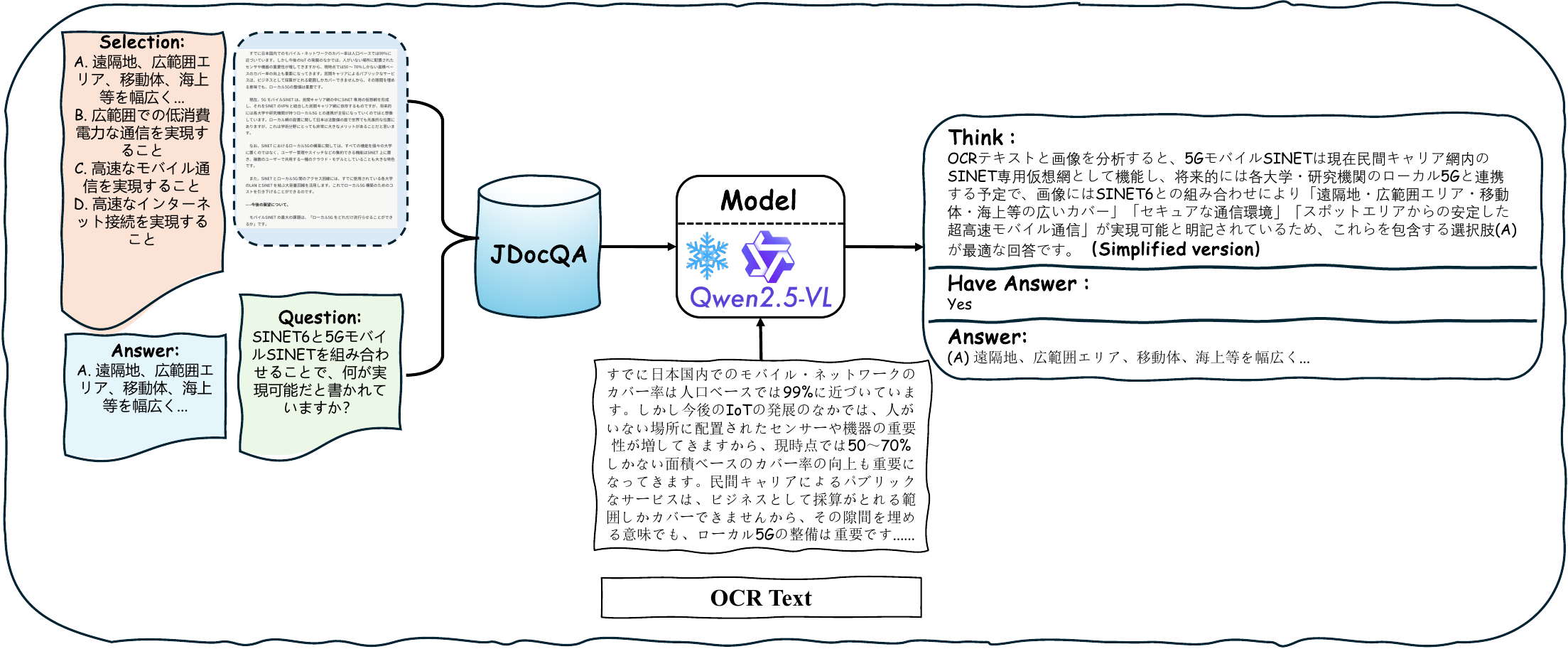}
  \caption{VLM Based Answer Validation System for OCR Text and Document Images}
  \label{fig:DataClean}
\end{figure}

\subsubsection{\textbf{Stage 2: Answer Feasibility Analysis}}

\begin{figure}[h]
  \centering
  \includegraphics[width=\linewidth]{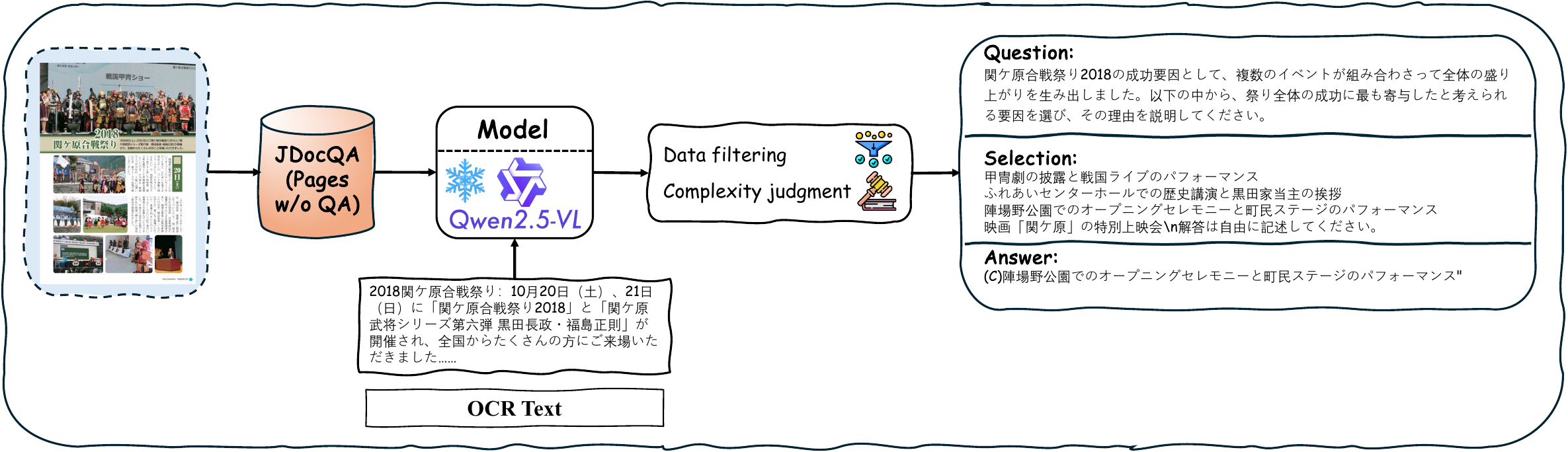}
  \caption{VLM Based Question-Answer Generation and Data Filtering System}
  \label{fig:Generate}
\end{figure}

To ensure training data quality, we implement a comprehensive feasibility analysis system that determines whether questions can be answered from available content through structured reasoning validation, as shown in Figure \ref{fig:DataClean}.

For each question-answer pair, the system constructs complete context including question, options, OCR text, and images. Through specially designed prompts, the model performs three-step analysis: detailed reasoning process, binary answerability judgment, and specific answer with supporting evidence. The system parses structured model outputs to extract three key components. The validation algorithm checks reasoning process logic, answer derivability, and evidence sufficiency. Only question-answer pairs passing all validation criteria are retained for training.

\subsection{Intelligent Data Augmentation}

Beyond filtering existing data, we address training data scarcity through systematic generation of high-quality reasoning questions targeting five distinct cognitive categories, as illustrated in Figure \ref{fig:Generate}.
The intelligent page selection algorithm first filters out non-content pages such as covers and table of contents, then scores pages based on content richness. The system prioritizes middle pages and ensures uniform sampling to guarantee comprehensive coverage. For selected pages, the generator creates specific generation templates according to five question types: comparative analysis, computational reasoning, conditional judgment, causal relationships, and comprehensive understanding. Each type employs different prompting strategies to ensure question reasoning complexity. Generated questions undergo multi-layer quality validation including length verification, complexity validation, answer support checking, option quality control, and deduplication mechanisms.

\subsection{Hybrid Text Retrieval Architecture}

\begin{figure*}[!h]
  \centering
  \includegraphics[width=\linewidth]{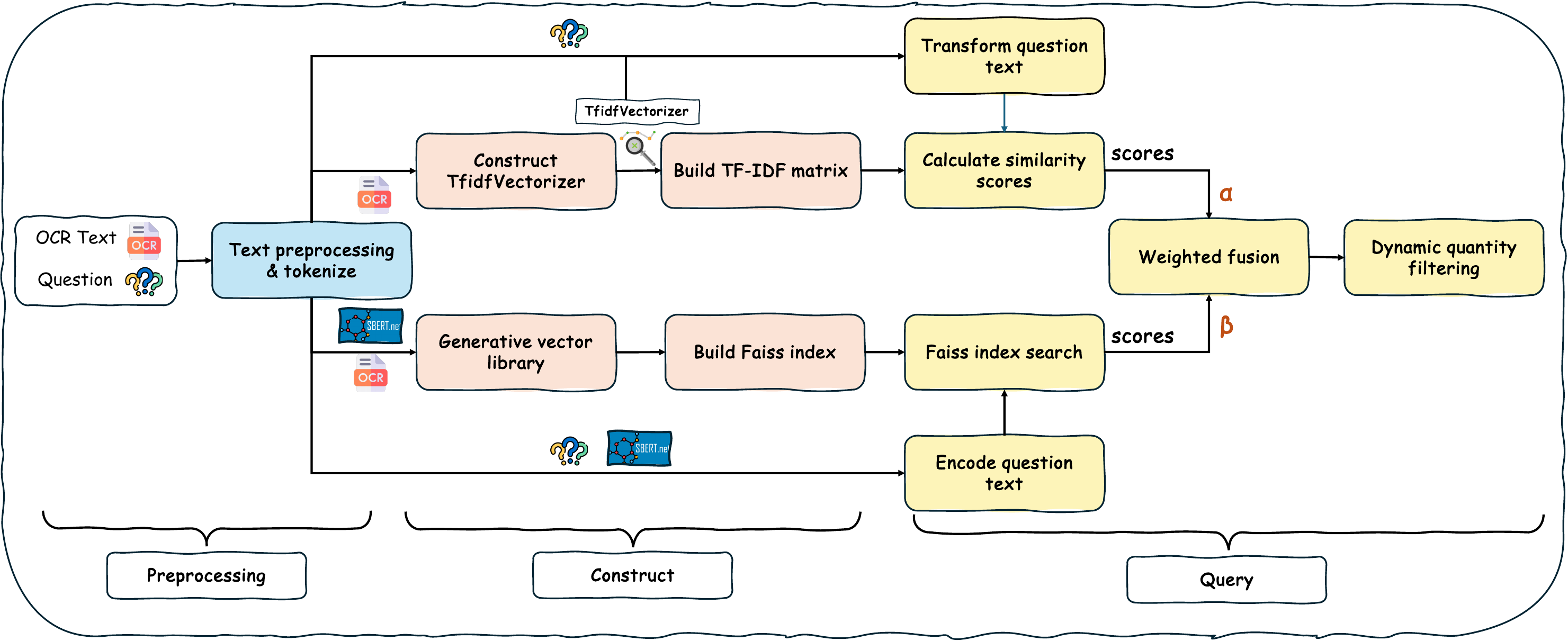}
  \caption{Overview of Retrieval Method}
  \label{fig:Retrieval}
\end{figure*}

\subsubsection{\textbf{Dynamic Ensemble Retrieval System}}
Our retrieval system combines complementary approaches through sophisticated fusion mechanisms that optimize both precision and recall while adapting to query characteristics, as illustrated in Figure \ref{fig:Retrieval}. The hybrid architecture addresses fundamental limitations of single-method retrieval by integrating lexical matching with semantic understanding, particularly crucial for Japanese text processing where morphological complexity and mixed script usage challenge traditional approaches.

\subsubsection{\textbf{Enhanced TF-IDF Lexical Retrieval}}
The TF-IDF component employs advanced tokenization strategies specifically optimized for Japanese text processing challenges. Traditional tokenization fails to handle the complexity of Japanese documents containing mixed scripts and technical terminology. Our enhanced approach implements a multi-stage tokenization pipeline that applies jieba-based morphological analysis for proper word segmentation, then employs regex-based fallback patterns to capture overlooked technical terms and numerical expressions.

The tokenization process begins with text normalization to standardize character encodings and remove formatting artifacts. This hybrid approach significantly improves term coverage and reduces information loss during preprocessing. TF-IDF configuration extends beyond standard parameters to capture Japanese linguistic patterns effectively, with n-gram ranges spanning from unigrams to 5-grams and maximum feature counts increased to 50,000 to accommodate extensive vocabulary diversity. Sublinear term frequency scaling with smoothed inverse document frequency prevents common particles from dominating relevance scores while preserving content-bearing terms' discriminative power.

\subsubsection{\textbf{Semantic Retrieval with Multilingual Embeddings}}
Semantic retrieval addresses TF-IDF's limitation in capturing paraphrased queries and conceptual relationships through dense vector representations. We employ multilingual-e5-large, a transformer-based model specifically designed for cross-lingual semantic understanding, which demonstrates superior performance on Japanese text compared to alternatives.

For processing mixed-language content, we employ the multilingual-e5-large~\cite{reimers-2020-multilingual-sentence-bert} model from the SentenceTransformers~\cite{reimers-2019-sentence-bert} library. Its sophisticated encoding pipeline includes text preprocessing steps like language detection and script normalization to ensure optimal encoding quality. The model generates 1024-dimensional dense vectors that capture semantic relationships beyond surface-level lexical similarity, enabling retrieval of conceptually relevant passages even when vocabulary differs significantly.

For efficient similarity computation across large document collections, we implement Faiss~\cite{douze2024faiss} indexing with approximate nearest neighbor search using IndexFlatIP with L2 normalization to compute cosine similarity:

\begin{equation}
\text{sim}(q, d) = \frac{\mathbf{v}_q \cdot \mathbf{v}_d}{|\mathbf{v}_q| \cdot |\mathbf{v}_d|}
\end{equation}

where $\mathbf{v}_q$ and $\mathbf{v}_d$ are the L2-normalized embedding vectors for query $q$ and document $d$ respectively, and $\cdot$ denotes the dot product operation. This formulation enables efficient similarity computation through inner product after normalization, leveraging Faiss's optimized IndexFlatIP implementation for scalable retrieval across large document collections.

\subsubsection{\textbf{Adaptive Result Selection Mechanism}}

The adaptive result selection mechanism dynamically determines retrieval quantity based on relevance confidence rather than fixed top-k values. The system employs a three-parameter framework to balance precision and recall while adapting to query complexity.

Results from TF-IDF and semantic retrieval are combined through weighted score fusion:
\begin{equation}
S_{final}(d) = \alpha \cdot S_{tfidf}(d) + \beta \cdot S_{semantic}(d)
\end{equation}

where $S_{final}(d)$ is the final relevance score for document $d$, $S_{tfidf}(d)$ and $S_{semantic}(d)$ are the normalized TF-IDF and semantic similarity scores respectively, and $\alpha, \beta$ are the weighting parameters with $\alpha + \beta = 1$.

The adaptive selection process is formalized as:
\begin{equation}
R = {d_i : (|R| < m) \lor (S_{final}(d_i) \geq \tau \land |R| < n)}
\end{equation}

where $R$ is the final result set, $m$ is the minimum result count ($top_m$), $n$ is the maximum result count ($top_n$), $\tau$ is the similarity threshold, and documents are ranked by $S_{final}$ in descending order.

This mechanism automatically adapts to query complexity variations, expanding retrieval for complex multi-part questions while constraining results for focused queries, ensuring high-quality results while preventing noise from low-relevance documents.

\subsection{Model Fine-tuning Strategy}
\begin{figure}[h]
  \centering
  \includegraphics[width=\linewidth]{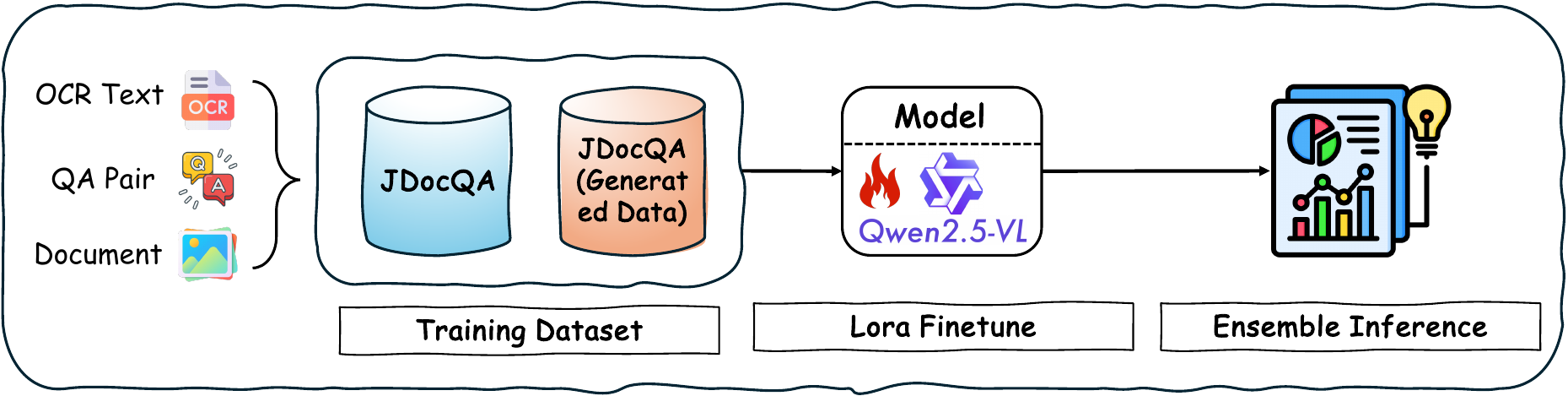}
  \caption{Model Fine-tuning and Inference Pipeline}
  \label{fig:Finetuning}
\end{figure}

\begin{figure}[h]
  \centering
  \includegraphics[width=\linewidth]{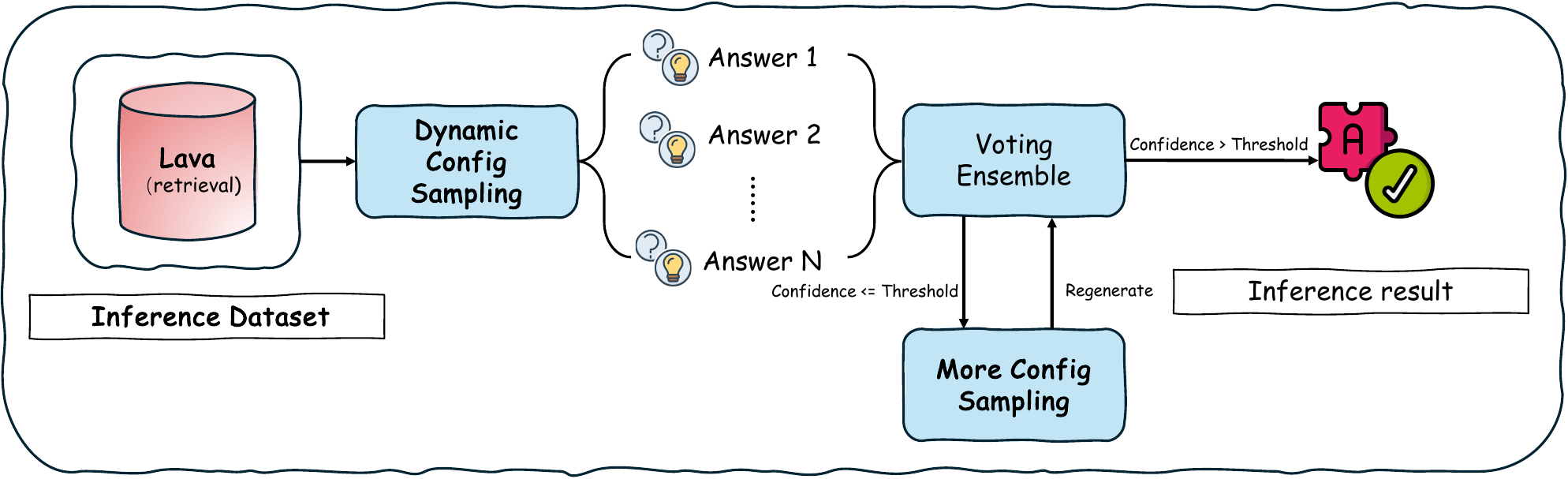}
  \caption{Model Ensemble Inference Pipeline}
  \label{fig:Infer_ens}
\end{figure}
Training data integration combines filtered original data, generated reasoning questions, and synthesized OCR content for augmentation. LoRA fine-tuning targets key model components with optimized adaptation parameters. The training process employs standard optimization techniques including cosine scheduling, gradient accumulation, and regularization, with performance monitoring on validation sets to guide convergence, as illustrated in Figure \ref{fig:Finetuning}.

\subsection{Ensemble Inference with Dynamic Configuration}
Our inference framework addresses prediction variability through a systematic ensemble approach that balances generation diversity with computational efficiency, as illustrated in Figure \ref{fig:Infer_ens}.

The ensemble strategy employs multiple generation configurations to capture different aspects of model knowledge. Deterministic configurations using greedy decoding provide stable baseline predictions, while probabilistic sampling configurations explore the model's uncertainty space through varied temperature, top-p, and top-k parameters. This diversified sampling ensures comprehensive coverage of potential response patterns while maintaining computational tractability.

To optimize efficiency, we implement an adaptive early stopping mechanism that monitors response convergence in real-time. The system tracks answer consistency across generated responses and terminates when confidence levels exceed threshold values, provided minimum response requirements are satisfied. This approach significantly reduces computational overhead while preserving ensemble quality.

The final prediction aggregation employs a multi-stage decision process. Answer choices are extracted through pattern matching and consensus is determined via majority voting. When multiple choices achieve equal support, selection criteria prioritize responses with superior structural clarity and content completeness, ensuring optimal answer quality for downstream evaluation.

\section{Experiment}
\subsection{Dataset}
\subsubsection{\textbf{JDocQA}}
The JDocQA dataset~\cite{onami2024jdocqa} comprises 11,600 Japanese document question-answering pairs spanning 5,504 PDF documents. To enhance text quality and consistency, we re-performed OCR text extraction on document images using deep learning models combined with data cleaning techniques. We merged the original training and validation sets (9,290 and 1,134 samples respectively) and applied rigorous filtering to ensure questions and answers can be logically derived from image content while eliminating unanswerable questions. This process yielded a four-option Japanese visual question answering dataset with 8,474 high-quality samples for training. Additionally, we implemented a VLM-based question-answer generation pipeline using Qwen2.5-VL to generate 1,808 additional high-quality QA pairs, bringing our total training dataset to 10,282 samples. The original test set contains 1,176 samples.
\subsubsection{\textbf{LAVA}}
The LAVA Challenge dataset \cite{lava2025challenge} originates from the ACM MM 2025 Workshop, containing 177 Japanese PDF document question-answering samples with a 10-choice format. Unlike traditional document QA tasks, this dataset does not provide specific page information, requiring comprehensive document understanding. We employ a Hybrid Text Retrieval Architecture combining TF-IDF vectorization (weight 0.6) and multilingual sentence embedding models (weight 0.4) for relevant page retrieval, with dynamic threshold filtering (minimum 3 pages, maximum 7 pages, similarity threshold 0.3) to optimize results.
\subsection{Implementation Details}

\subsubsection{\textbf{Training}}
This experiment employs LoRA (Low-Rank Adaptation) fine-tuning to train the Qwen2.5-VL-72B-Instruct model. Training is conducted on 8*A100 (80GB) GPUs using bfloat16 precision for improved training efficiency. LoRA parameters are set to rank=32 and alpha=64. For training hyperparameters, the learning rate is set to 1e-4 with 7 epochs of training, per-device batch size of 1, and gradient accumulation steps of 4, resulting in an effective batch size of 32. The maximum sequence length is limited to 16384 tokens with a maximum generation length of 256 tokens. The optimization strategy incorporates a 5\% warmup ratio and utilizes DeepSpeed ZeRO-3 for memory optimization. Model checkpoints are saved every 200 steps with validation performed every 50 steps. The entire training process takes approximately 24 hours.

\subsubsection{\textbf{Inference}}
The inference stage adopts an ensemble learning strategy to enhance model performance. The system deploys the fine-tuned model using the LMDeploy framework across 8 GPUs, performing ensemble inference with 20 dynamically generated decoding configurations, including greedy decoding, sampling strategies with varying temperature parameters (0.1-1.5), and diverse top-p and top-k combinations. To improve efficiency, an early stopping mechanism is implemented: inference terminates when confidence reaches 0.8 with at least 10 completed inferences. The final answer is determined through a voting mechanism, selecting the most frequently occurring option as the prediction.

\begin{table}
  \caption{Quantitative comparison of different models on JDocQA}
  \label{tab:JDocQA}
  \begin{tabular}{cccc}
    \toprule
    Model&Y/N&Fact.&Num\\
    \midrule
    InstBLIP (blank)& 65.58 & 16.27 & 19.88 \\
    InstBLIP (img) & 68.63 & 15.34 & 19.88\\
    InstBLIP (bbox)& 72.78 & 18.13 & 19.29 \\
    Qwen2.5-VL-72B & \textbf{83.04} & \textbf{52.66} & \textbf{44.12} \\
  \bottomrule
\end{tabular}
\end{table}

\begin{table}
  \caption{Quantitative comparison of different models on LAVA}
  \label{tab:LAVA}
  \begin{tabular}{ccc}
    \toprule
    Model&Accuracy(\%)\\
    \midrule
    Qwen2.5-VL-7B & 37\\
    Qwen2.5-VL-32B & 51\\
    Qwen2.5-VL-72B & \textbf{59} \\
    InternVL3-78B & 49 \\

  \bottomrule
\end{tabular}
\end{table}

\begin{table}
  \caption{Ablation Study on LAVA with Qwen2.5-VL-72B}
  \label{tab:Ablation}
  \begin{tabular}{cccc}
    \toprule
    Retrieval & Train(Lora)  & Ensemble Inference & Accuracy(\%)\\
    \midrule
     $\times$ & $\times$ & $\times$ & \textbf{OOM}\\
     $\checkmark$ & $\times$ & $\times$ & 45\\
     $\checkmark$ & $\checkmark$ & $\times$ & 54\\
     $\checkmark$ & $\checkmark$ & $\checkmark$ & \textbf{59}\\
    \bottomrule
  \end{tabular}
\end{table}
\subsection{Result Comparison}

Table \ref{tab:JDocQA}  presents results across different question types on JDocQA. Our approach using Qwen2.5-VL-72B significantly outperforms InstBLIP baselines across all categories. For Yes/No questions, we achieve 83.04\% accuracy compared to InstBLIP's best result of 72.78\%. The improvement is more substantial for factual questions (52.66\% vs. 18.13\%) and numerical questions (44.12\% vs. 19.88\%), demonstrating the framework's effectiveness in complex reasoning tasks.

Table \ref{tab:LAVA} shows the performance comparison of different vision-language models on the LAVA dataset. Our framework achieves the best results with Qwen2.5-VL-72B, reaching 59\% accuracy. The results demonstrate that larger models generally perform better, with the 72B model outperforming the 7B (37\%) and 32B (51\%) variants. Notably, our framework with Qwen2.5-VL-72B also surpasses InternVL3-78B (49\%), indicating the effectiveness of our methodological approach.

Table \ref{tab:Ablation} validates the contribution of each framework component on the LAVA dataset. Without retrieval mechanisms, we input all images extracted from the PDF directly into the large model. However, due to the large number of images in typical PDFs, this leads to an explosion in the number of vision tokens, quickly exceeding the model's token capacity and resulting in out-of-memory (OOM) errors. This highlights the necessity of our hybrid retrieval architecture to ensure computational feasibility. The retrieval component alone enables 45\% accuracy by efficiently narrowing down relevant content, LoRA fine-tuning further improves performance to 54\%, and ensemble inference achieves the final 59\% accuracy. Each component contributes meaningfully, with retrieval being essential for tractability and fine-tuning providing the largest performance gain.

\section{Conclusion}
We present AdaDocVQA, a unified framework for document visual question answering that addresses critical challenges in processing long documents and low-resource languages. Our approach combines hybrid text retrieval, intelligent data augmentation, and adaptive ensemble inference to achieve robust performance in resource-constrained scenarios.

Extensive experiments on Japanese document VQA benchmarks validate our framework's effectiveness. On JDocQA, we achieve substantial improvements across all question categories, with 83.04\% accuracy on Yes/No questions, 52.66\% on factual questions, and 44.12\% on numerical questions. On the LAVA dataset, our method attains 59\% accuracy, outperforming larger competing models. Ablation studies demonstrate that each component contributes meaningfully to overall performance.

The framework makes three primary contributions: a hybrid retrieval architecture optimized for Japanese text processing, an automated data generation pipeline with multi-level quality control, and an adaptive ensemble strategy that balances prediction robustness with computational efficiency. Our work establishes new state-of-the-art results for Japanese document VQA while providing a scalable foundation for other low-resource languages and specialized domains.

Future research directions include extending the framework to additional languages and exploring more sophisticated retrieval mechanisms for extremely long document collections.

\bibliographystyle{ACM-Reference-Format}
\bibliography{sample-base}

\end{document}